\documentclass[letterpaper, 10 pt, conference]{ieeeconf}  
\IEEEoverridecommandlockouts                              
\usepackage{graphics} 
\usepackage{epsfig} 
\usepackage{mathptmx} 
\usepackage{times} 
\usepackage{amsmath} 
\usepackage{amssymb}  
\usepackage{hyperref}
\usepackage{url}
\usepackage{multirow}
\usepackage{gensymb}
\usepackage[utf8x]{inputenc}
\usepackage{CJK}
\hypersetup{
    colorlinks=true,
    linkcolor=blue,
    filecolor=magenta,      
    urlcolor=cyan,
}

\title{\LARGE \bf
A Collaborative Visual SLAM Framework for Service Robots
}

\author{Ming Ouyang, Xuesong Shi, Yujie Wang, Yuxin Tian, \\
Yingzhe Shen, Dawei Wang, Peng Wang, Zhiqiang Cao
\thanks{M. Ouyang and Z. Q. Cao are with Institute of Automation, Chinese Academy of Sciences, Beijing, China, and School of Artificial Intelligence, University of Chinese Academy of Sciences, Beijing, China.}%
\thanks{X. Shi, Y. Wang, Y. Tian, Y. Shen, D. Wang and P. Wang are with Intel Corporation, Beijing, China.}%
\thanks{Correspondence to xuesong.shi@intel.com.}
}

\begin{document}
\begin{CJK*}{GBK}{song} 
\maketitle

\begin{abstract}

We present a collaborative visual simultaneous localization and mapping (SLAM) framework for service robots. With an edge server maintaining a map database and performing global optimization, each robot can register to an existing map, update the map, or build new maps, all with a unified interface and low computation and memory cost. We design an elegant communication pipeline to enable real-time information sharing between robots. With a novel landmark organization and retrieval method on the server, each robot can acquire landmarks predicted to be in its view, to augment its local map. The framework is general enough to support both RGB-D and monocular cameras, as well as robots with multiple cameras, taking the rigid constraints between cameras into consideration. The proposed framework has been fully implemented and verified with public datasets and live experiments.

\end{abstract}

\section{Introduction}

Recently, we see a rapidly increasing number of service robots being deployed in various public places, such as hospitals, hotels, restaurants and malls, for delivery, cleaning, disinfection and inspection, etc. Most robots are expected to be able to move autonomously, so generally they would need a pre-built or evolving map of the place of interest for localization and planning. An evolving map is preferred, since most service robots work in dynamic and daily changing environments. For a single robot, its localization and map updating can be performed with a Simultaneous Localization And Mapping (SLAM) system \cite{cadena2016past}. However, for multiple robots serving in one place, as each may be turned on and off from time to time, ideally there shall be a centralized map database of this place that is always up-to-date and can be used by any robots registered to this place at any time. The map database can be maintained by an on-premise edge server \cite{shi2016edge}. While edge server-assisted SLAM system has been proposed recently \cite{ali2020edge}, there lacks a framework to allow multiple robots to simultaneously work within a centralized map database on the server.



Collaborative SLAM algorithms aim to enable multiple independently moving agents to exchange spatial information for better robustness, efficiency and accuracy in localization and mapping \cite{zou2019collaborative}. Many recent works adopt a client-server architecture \cite{forster2013collaborative}\cite{riazuelo2014c2tam}\cite{morrison2016moarslam}\cite{schmuck2017multi}\cite{schmuck2019ccm}. Although they are originally designed for Augmented Reality (AR) or Micro Aerial Vehicle (MAV) applications, many of the techniques can be applied to service robot scenarios. In this paper, we give a careful review of the literature, and present a multi-agent visual SLAM system dedicated for service robot applications. We elaborate our design with the hope of pushing forward a standard interface for different kinds of service robots to benefit from a unified SLAM service powered by edge servers. We can imagine many open problems, such as privacy preserving and security assurance, but we believe that this is the right direction to build smart spaces where people can be better served.


The first problem in designing a multi-agent SLAM system is the partition of modules between the server and clients, to balance between a powerful centralized system and fully autonomous agents. For the targeted service robot scenarios, we assume that an edge server is always on-site with good (but non-perfect) communications with each client. Therefore, the maps are mainly maintained by the server, while each client keeps only a small local map for real-time pose tracking. The clients also optimize their local maps with new observations, so that they can retain autonomy even if the communication is temporarily unavailable. Instead of sending images to the server, each client reports only keyframe features and optimized landmarks. It reduces both bandwidth usage and computation overhead at server side for each robot, making the design scalable for supporting many robots.

For each client to benefit from others' and previous observations, we propose a novel method for the clients to efficiently query the map database from the server. Instead of downloading the full map, only landmarks within the client's current view are retrieved. This is efficiently performed by a grid map-based landmark searching algorithm. The retrieved landmarks may originate from other clients' observations, or other sessions. For each new keyframe, the server will try to detect loops with existing keyframes and perform a map merging or map optimization whenever possible.

A service robot can equip multiple cameras to enlarge its view for more robustness in narrow or crowded places. The proposed framework can naturally accommodate this setting by treating each camera as a separate client and adding inter-camera constraints for global optimization at the server side. The system can keep working when some of the cameras temporarily fail.

In summary, the contributions of this work include
\begin{itemize}
    \item a collaborative visual SLAM framework with an efficient communication pipeline;
    \item a novel landmark organization and retrieval method to allow real-time information sharing;
    \item a fully-implemented and verified system that can work with many robots, each with one or more cameras.
\end{itemize}

\section{Related Works}

\subsection{Visual SLAM}

Visual SLAM algorithms, by the kind of visual information they use, can be categorized into direct methods and feature-based ones. The former uses pixel intensities in the images to estimate camera motion and environmental structures by minimizing a photometric error, and can generate a dense or semi-dense reconstruction of the scene. Feature-based ones, despite giving only a sparse map, are generally more accurate and robust for that they select reliable features in the scene, such as keypoints, lines, planes or objects. They are also more flexible as the features from different trajectories can be matched and optimized together, which is crucial for generalizing to a multi-session or multi-agent setting. Representative works of keypoint and optimization based SLAM systems include PTAM \cite{klein2007parallel}, ORB-SLAM \cite{mur2015orb}, ORB-SLAM2 \cite{mur2017orb} and OpenVSLAM \cite{sumikura2019openvslam}. Our system in this work is developed based on OpenVSLAM.

\subsection{Multi-agent SLAM}

The motivation of multi-agent SLAM is to allow multiple devices to build and re-use maps. Castle et al. \cite{castle2008video} extends PTAM to a multi-camera multi-map setting, showing that multiple trackers can work within the same map, and each tracker can be efficiently re-localized on multiple maps. There is no map merging mechanism because constructing a global map is not mandatory in the targeted wearable AR applications. In robotic applications, however, maps with overlapped areas are preferred to be merged together, so as to support robot autonomy over a large area. Forster et al. \cite{forster2013collaborative} propose Collaborative Structure from Motion (CSfM) for MAVs. Each MAV runs a visual-inertial odometry system, and sends image features and keyframe poses to a base station for centralized mapping. A more recent work is MOARSLAM \cite{morrison2016moarslam}, where maps are constructed by each agent, but can be synchronized with a remote server. The server connects two maps from different agents once a loop between them is reliably detected.


Collaborative localization is also considered in the literature. A seminal work is CoSLAM \cite{zou2012coslam}, which presents several ways to enhance localization by utilizing common observations among multiple independently moving cameras. In particular, cameras with overlapped views are dynamically grouped, and inter-camera pose estimation and mapping can be performed within the same group. However, despite the merits, such methods are less re-used in later works because they not only require all the cameras to be synchronized, but also imply a tightly coupled pipeline between agents. Instead, the client-server architecture has attracted more attention. It is enhanced to support not only collaborative mapping, but also collaborative localization by real-time map synchronization between the server and clients \cite{riazuelo2014c2tam}\cite{schmuck2017multi}\cite{schmuck2019ccm}.

The Cloud framework for Cooperative Tracking And Mapping (C$^2$TAM) \cite{riazuelo2014c2tam} is a multi-agent SLAM system with a modern client-server infrastructure. It partitions the modules of PTAM into server-side and client-side. The major consideration is to leverage the storage and computational resources in the cloud server while maintaining the real-time pose estimation capability at client side. To this end, the client performs only tracking and relocation (local re-localization), while the server provide mapping, place recognition (global re-localization) and map fusion as services. To save bandwidth, each client sends only keyframes to the server. However, the server has to send updated map to the client every time the map is optimized, which can be costly when the map gets large.

In \cite{schmuck2017multi} and CCM-SLAM \cite{schmuck2019ccm}, each client runs tracking and local mapping, maintaining a local map consisting of a fixed number of nearest keyframes and related landmarks. The local map is synchronized with the server, allowing bi-directional map updates, thus enabling indirect information sharing between agents through the server.

Our system has a similar architecture with CCM-SLAM, but has been particularly optimized for ground robots. We propose an efficient grid map-based landmark searching method using ground plane. Unlike CCM-SLAM with monocular devices, we support both monocular and RGB-D cameras. We also show the use case of robust visual SLAM with multiple heterogeneous sensors on one robot.

Besides these multi-agent SLAM algorithms with a base station or a cloud server, there are also works on decentralized multi-agent SLAM algorithms. In such cases the agents leverage local communication to reach consensus on a common estimate. Such kind of algorithms are generally more costly in computation and communication, and more fragile to spurious measurements and matches as each agent only has access to partial and local information \cite{cadena2016past}. Thus decentralized SLAM algorithms are less practical for service robot scenarios, in which high reliability is required and server deployment is often possible.

\begin{figure*}[t]
\centering
  \vspace{-0in}
  \hspace*{-1.35in}
  \includegraphics[width=7.5in]{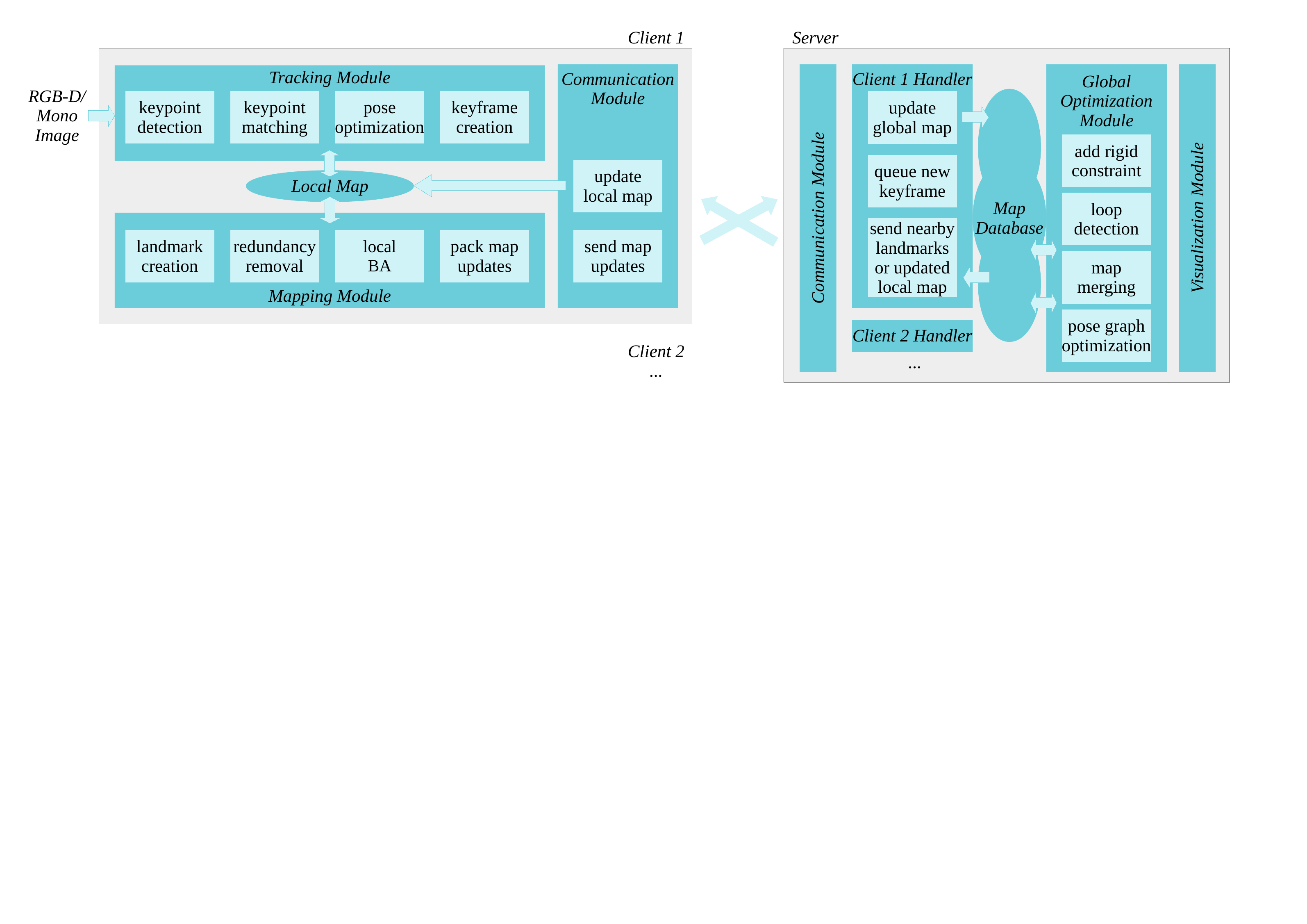}
  \hspace*{-1.6in}
  \vspace{-3.1in}
  \caption{The proposed collaborative visual SLAM framework features a server to maintain all the maps. Each client performs visual odometry with a small local map, which is synchronized with the server's map database in real-time.}
  \vspace{-0.5cm}
  \label{fig_system}
\end{figure*}

\subsection{Multi-camera SLAM}

SLAM with multiple cameras mounted on one rigid body can be viewed as a special case of multi-agent SLAM. For a pair of synchronized and calibrated cameras with largely overlapped views, they are usually considered as one stereo camera setup, where frames from one camera are tracked and mapped, while frames from the other help triangulate the keypoints \cite{mur2017orb}. With more cameras on one device, its field of view can be greatly enlarged, thus better mapping efficiency and localization robustness. Kaess and Dellaert \cite{kaess2006visual} design a rig with 8 cameras distributed equally along a circle, and show that it can efficiently and accurately map a large scene.

For a general multi-camera setting, Kuo et al. \cite{kuo2020redesigning} presents several techniques to build a more adaptive SLAM system with fewer hyper-parameters. One of the techniques is to organize all the landmarks in a voxel grid, so that for an arbitary camera pose, landmarks within its view can be efficiently retrieved by sampling the camera frustum, rather than relying on keyframe covisibility which has several drawbacks \cite{muglikar2020voxel}. We take a step further from this work and propose a grid map based landmark retrieval method that are efficient for collaborative service robot scenarios.


\subsection{Multi-session SLAM}

Multi-session SLAM, or lifelong SLAM in some contexts, aims to reuse previous map information for later trajectories in the same place. It can be viewed as a sub-topic of multi-agent SLAM, and the proposed methods can usually be directly incorporated into the server side of a client-server architecture. The concept of {\it Atlas} is introduced in \cite{bosse2004simultaneous}, showing that a global topological map can be efficiently built and reused by multiple local SLAM sessions. Several open-source SLAM systems support inter-session loop detection (also known as place recognition) and map merging, including Maplab \cite{schneider2018maplab}, RTABMap \cite{labbe2019rtab} and ORB-SLAM3 \cite{campos2020orb}. For place recognition, recent works show that deep convolutional neural network based feature extraction can greatly outperform conventional methods and can be well integrated into keypoint-based SLAM pipelines \cite{li2020dxslam}. 



\section{Multi-Agent SLAM System}


Our goal is to build a ready-to-use visual SLAM frame for multiple ground robots working in one place. In particular, the framework should 
\begin{itemize}
    \item allow multiple robots to build and reuse one or more global maps of the place, without any assumption of each robot's movement,
    \item enable timely information sharing among robots to collaboratively address scene changes and expansions,
    \item allow robots with multiple cameras to take full advantage of the overall field of view,
\end{itemize}
with the constraints of a) no hardware synchronization of cameras, b) bounded on-board computation and memory cost for each robot, and c) non-perfect communication channels with possible message loss.

For this goal, it is natural to employ an on-premise edge server to manage the global maps and communicate with each robot via wireless connections. Instead of treating each robot as a client, we choose to make each camera (or camera set for stereo or RGB-D sensors) to be a client to communicate with the server individually. Thus, a robot equipped with multiple cameras may have multiple client-side algorithm running on-board. Then if needed, a separate fusion module can be designed to give the final localization of this robot by merging results from these clients with a filter and a policy to deal with individual tracking failures.

\subsection{System Design}

Our system has a modular design, as shown in Fig.~\ref{fig_system}. In this subsection, we briefly explain each module's functionality and the reason for them to be in the client or server side.

The \textit{tracking module} estimates the pose of each frame by matching the extracted keypoints to the local map. It should be running at client side to ensure real-time response.

The \textit{mapping module}, taking care of keyframe creation, landmark creation, and local bundle adjustment, should also be running at client side. Otherwise if each client relies on the server to update the map, its autonomy would be greatly constrained by the communication.

The \textit{global optimization module} performs loop detection and map optimization. These features have no real-time requirement and may require a large memory, making it reasonable to run on the server. Besides the common loop closures within one trajectory, it also detects inter-client and inter-session loops, and merges corresponding maps whenever possible. This module can also deal with the prior rigid constraints between cameras on the same robot. Because map optimization occurs much more frequently in a multi-robot system, we use an efficient hierarchical optimization algorithm \cite{tian2021hierarchical}.

The \textit{map database} at server side stores the maps for all clients. Each map consists of keyframes and landmarks. Each keyframe or landmark has a unique ID, and can be efficiently retrieved by their ID based on a hash table index.

The \textit{communication modules} connect the server and each client. At either side, it transmits messages queued by other modules, and invokes the proper callback when receiving messages. All the messages are in a unified format, containing a variable number of keyframes and landmarks.

For each alive client, there is a \textit{client handler} at server side to process messages from this client, fuse the information into the map database, and send back messages to the client with nearby landmarks or updated local map.

With this modular design, the framework can be implemented based on any keypoint and keyframe-based SLAM system. Our implementation is extended from OpenVSLAM \cite{sumikura2019openvslam}, re-using its tracking, local mapping, and pose graph optimization modules. The system supports monocular, stereo and RGB-D cameras. Unlike most existing collaborative SLAM works which require the same sensor setup, our framework naturally supports robots with different cameras.

The communications between clients and server are based on an asynchronous message delivering mechanism with a unified message format. Each message has a keyframe array and a landmark array, and a few metadata and control commands.


\subsection{Collaborative Mapping}

When a client starts, or recovers from a tracking failure, it initializes a new session and builds a new local map. The map is synchronized with the server's map database from then on. The client will assign a unique ID to each map element (keyframes and landmarks) upon their generation. The server will assign a map ID whenever it receives a message from a new session.

\textit{Map updates from clients to server -} When a client generates a new keyframe, it sends a message to the server after local mapping. The message contains the keyframes and landmarks that are newly generated, updated, or pruned. If a keyframe is new, its keypoints will be sent; otherwise only pose estimates and other variables are included. For each landmark, its location, observations and descriptor are included, as all these information can be updated. Because maps are maintained by the server, the client keeps only a small set of keyframes and landmarks which are needed for the local bundle adjustment, and will remove previous ones out of its memory. So the client's memory usage can be low and bounded even in a long session.

\textit{Map merging -} When a client handler receives a newly generated keyframe, after adding it into the map database, it puts the keyframe ID into a queue shared between all the handlers. The global optimization module retrieves each new keyframe from this queue and performs a loop detection with all existing keyframes in the database. If a loop is found and verified between two keyframes in different maps, the smaller map will be merged into the larger one, by updating the map ID and world pose for the keyframes and landmarks in the former. A pose graph optimization is performed for each loop closure. Because the optimization is time-consuming, the message processing of client handlers working on the related map(s) is paused during optimization. After that, the handlers will process the accumulated messages with aligning the world poses onto the updated map coordinates. They also help update the clients' local maps by sending a message containing updated keyframes and landmarks within their client's local map.

With this design, the system naturally supports intra-session and inter-session loop closures, as well as re-localization, simply by merging a newly generated map into an existing one. An exception is initialization with monocular cameras. In this case, an explicit place recognition request containing the current frame features will be sent to the server, and the client will not start tracking until getting successful initial pose estimate and nearby landmarks from the server. The reason for this choice is that the initialization of monocular SLAM can be difficult and time-consuming, and so can be the merging of maps without an absolute scale. With our choice, the initialization of monocular clients is fast and reliable if it is on a mapped area, which we assume should often be the case for service robot applications\footnote{RGB-D and stereo cameras are usually preferred in service robot design, while monocular cameras can be used as auxiliary cameras or by small-sized devices to track pose with a known map.}. If the monocular client then explores into a new area, it will continue tracking and extend the global map, as it has already been working with the correct map scale.


\subsection{Collaborative Localization}

\begin{figure}[t]
\centering
  \vspace{-0in}
  \hspace{0in}
  \includegraphics[width=2.7in]{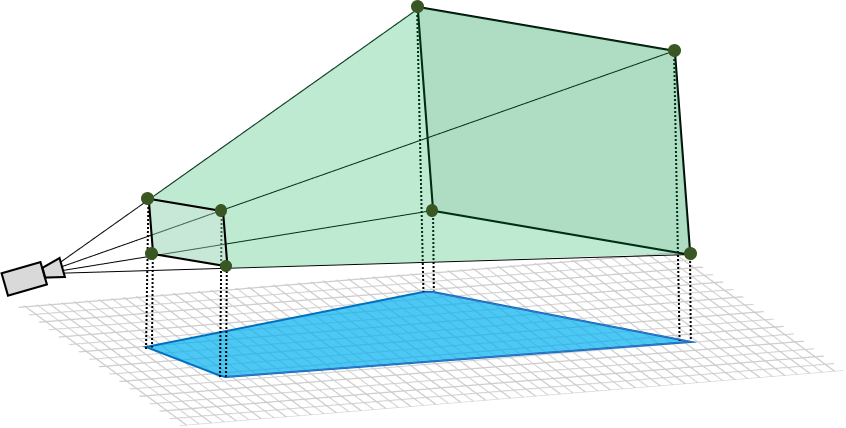}
  \hspace{0in}
  \vspace{-0in}
  \caption{The grid map-based method to retrieve nearby landmarks for a given camera pose. All the landmarks are indexed by a 2D grid (grey) which is coarsely aligned with the ground plane. Given a camera pose and intrinsics, we project the eight vertices (green dots) of its frustum (green area) onto the grid, calculate the grid cells that fall into the convex hull of the projected area (blue), and retrieve landmarks within these cells with additional checking.}
  \vspace{-0.5cm}
  \label{fig_grid_map}
\end{figure}

A core problem in collaborative SLAM is to share spatial information between clients, so that a client's localization can benefit from observations by other clients, or those from previous sessions, or those from the same session but had been shifted out from the local map. In a client-server system, this can be done by server sending the global map to each client \cite{riazuelo2014c2tam}, or more efficiently, sending only selected keyframes and landmarks that are close to the client's current keyframe \cite{schmuck2019ccm}. However, finding close keyframes and landmarks is non-trival. CCM-SLAM \cite{schmuck2019ccm} retrieves keyframes with the strongest covisibility, which is a powerful heuristic, but would fail to retrieve nearby landmarks observed from a largely different view angle. A recent work \cite{muglikar2020voxel} uses a voxel map index to efficiently retrieve landmarks within a given camera view, but only a subset of them, as the method requires sampling on the image plane and rays.

\textit{Efficient landmark retrieval -} We design a method for the server to efficiently retrieve all the landmarks within a given camera view. For each map, a 2D grid map index is implicitly built by discretizing the x- and y-coordinates of each landmark. A hash table indices all the landmarks with their grid coordinates as the key, so that given a map coordinate, the landmarks in the corresponding grid cell can be efficiently retrieved. When a camera pose is given, we calculate the map position of the eight vertices of its view frustum, as illustrated in Fig. \ref{fig_grid_map}. A 2D convex hull is constructed with the x- and y-coordinates of the eight points. Then for all the grid cells within the axis-aligned bounding box of the hull, a 2D point-polygon test is performed to check if the cell center lies within the hull. If yes, then the landmarks in this cell will be retrieved, and a re-projection check is performed for each landmark to filter out those outside the camera's view.


A 2D grid map is usually sufficient to index landmarks for ground robots, when landmarks distribute on a large scale on the two axes parallel with the ground while clustered on the height direction. To make sure that the x-y plane of the map coordinates is coarsely aligned to the ground, we define the map origin to be the robot base pose when initializing each session\footnote{The transform from the camera frame to the base frame is usually available for service robots.}. Nevertheless, our landmark retrieval method can be directly extended to a 3D grid if needed.


\textit{Local map augmentation -} Whenever the server receives a new keyframe, the corresponding client handler will retrieve all the landmarks within the view of this keyframe, and send them to the client to augment its local map. To save bandwidth, the landmarks that are lastly updated by the same client are excluded from the message. The client matches the received landmarks to the keypoints of its subsequent frames, and uses the matches in pose estimation and local mapping. In this way, not only can all the clients benefit from existing landmarks in mapped area, but also the map drift can be mitigated in long-term mapping.

\subsection{Rigid Constraints for Multi-camera Robots}

Service robots often face challenging visual conditions like featureless walls or serious occlusion, where visual SLAM is likely to fail \cite{shi2020we}. This can be mitigated by equipping multiple cameras with different views, e.g. one front-facing and one rear-facing. In such cases, the rigid constraint between cameras shall be considered in order to a) improve accuracy when multiple cameras works and b) keep tracking in the current map whenever at least one camera works. We achieve both goals by treating each camera as an independent client and fusing their maps at the server side by connecting concurrent keyframes from any pair of clients that have a known rigid constraint. Whenever one client has lost tracking, it can easily get back into the previous map when re-initialized, with the help of the rigid constraints with other cameras and map merging.

Because the cameras are not hardware synchronized and the keyframe selection is independent between clients, it may be rare that two keyframes are generated at a very close time. To relax this condition, we generate a virtual keyframe for one of the clients with the same timestamp of the other client's keyframe. The pose of this virtual keyframe is interpolated between the two closest keyframes. We then add the pre-defined constraint to the virtual keyframe and the keyframe from the other client.

If a constraint is added for two keyframes on different maps, a map merging operation will be triggered. Otherwise, it may trigger a pose graph optimization, in the same way as loop closure does. To avoid frequent optimizations, they are triggered only if the relative pose between the two keyframes diverges from the constraint to a certain threshold.

Because the clients for multi-cameras are loosely coupled, the design naturally supports heterogeneous multi-cameras, such as one RGB-D camera and several auxiliary monocular cameras to ensure a large overall fields of view for robust localization in complex environments.

\section{Experimental Results}

\begin{figure*}[t]
\centering
  \vspace{-0in}
  \centering
  \includegraphics[scale=0.5]{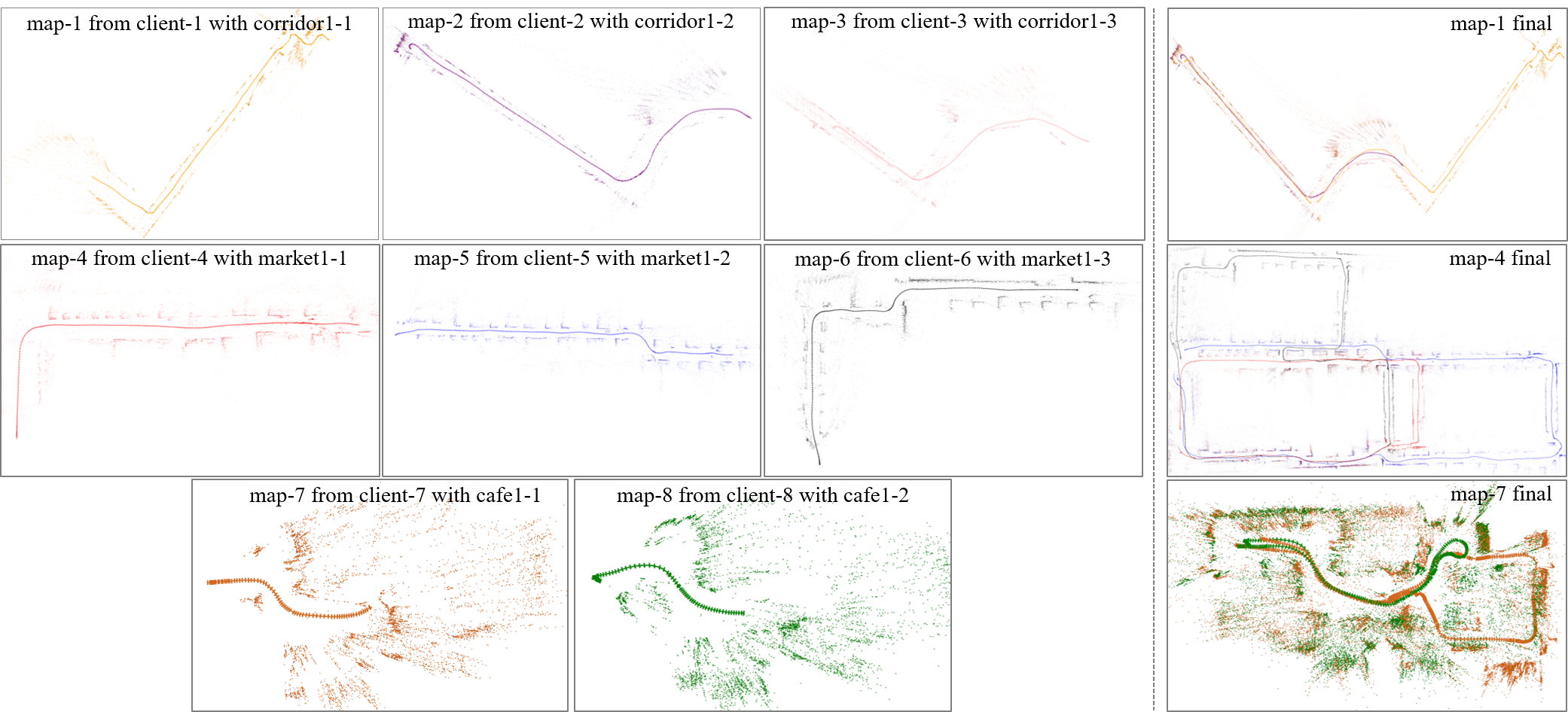}
  \caption{The visualized keyframes and landmarks at server side when it simultaneously handles 8 clients with different RGB-D sequences from the OpenLORIS-Scene datasets. Left: at an early time, 8 maps are built. Right: at a later time, they have been merged into 3 major maps. Better view in color.}
  \label{fig_openloris}
\end{figure*}


\begin{figure}[!ht]
  \centering
  \includegraphics[width=3.4in]{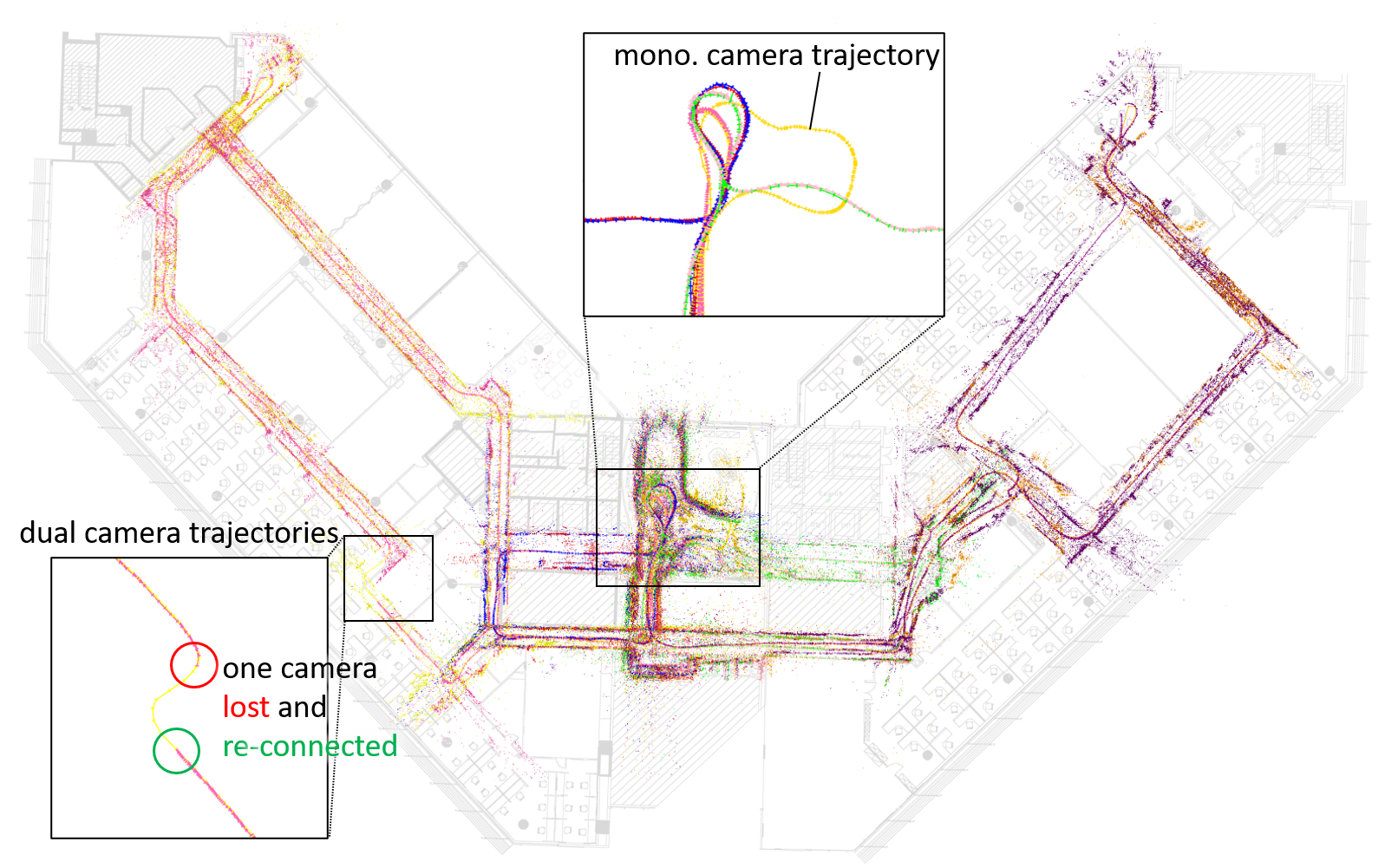}
  \caption{The colored trajectories and points represent a map of an office area (approx. 90 x 60 meters), overlayed on its floorplan. The map is built with a dual-RGB-D camera robot, with 4 dual-RGB-D sessions and 1 monocular camera session (9 clients in total, plotted in different colors). The subplot in the lower left shows zoomed-in trajectories where one camera (red traj.) gets lost (red circle) and then gets back to the map (green circle) with the rigid constraint from the other camera (light yellow traj.). The subplot in the top shows part of the monocular camera's trajectory (yellow).}
  \label{fig_ryc}
\end{figure}


The proposed system is verified with both public data and live experiments. The public data are from the OpenLORIS-Scene datasets \cite{shi2020we}, which are collected with a real service robot in typical indoor scenes.

\subsection{Multi-agent Multi-map Collaborative SLAM}

We test the system with eight RGB-D data sequences from the OpenLORIS-Scene datasets. They are from the largest two scenes, {\verb market } and {\verb corridor }, and a smaller one, {\verb cafe }. In our test, the eight sequences are concurrently played in real time and subscribed by eight clients to simulate collaborative mapping of eight robots. The clients are distributed on three machines: an Intel NUC mini-PC with i7-8809G, a desktop with i7-7820X, and a Dell laptop with i5-6300HQ. The server runs on another Intel NUC. The four machines are connected with an office network. Fig. \ref{fig_openloris} shows the intermediate and final maps at the server side, where keyframes and landmarks are visualized. While each client initialized with a separate map, those in the same scene eventually merged together. There are several more maps of {\verb corridor } in the final state that are not shown in the figure. The reason is that some clients have built multiple maps after tracking failures, and some of the maps are not looped with the largest map due to significant viewpoint changes. The server's CPU usage during the test is 100\%-150\% when handling all eight clients, indicating that this quad-core machine can support even more clients.

We compare the localization accuracy with OpenVSLAM \cite{sumikura2019openvslam}, as we have largely re-used its implementation of visual odometry. OpenVSLAM does not support map merging, so it is evaluated with each sequence. We test with the {\verb market } sequences because they are the only ones that can be fully tracked. Table \ref{tab_accuracy} shows the results, where numbers are overall comparative. Our system gives slightly better accuracy for {\verb market1-1 } and {\verb market1-2 }, most likely because that they do not have a self-loop to mitigate drifts but can benefit from inter-session loops in our system. 



\begin{table}[tp]
\begin{center}
\caption{Absolute Trajectory Error (ATE) of the Final Pose Estimation (in Meters)}
\label{tab_accuracy}
\begin{tabular}{c|c|c|c}
\hline
 & market1-1 & market1-2 & market1-3\\
\hline
OpenVSLAM \cite{sumikura2019openvslam} & 2.53 & 3.70 & \textbf{2.21}\\
\hline
Collaborative SLAM (ours) & \textbf{2.51} & \textbf{3.34} & 2.45\\
\hline
\end{tabular}
\end{center}
\end{table}

\begin{table}[tp]
\begin{center}
\caption{Time Cost of Major Procedures}
\label{tab_time}
\begin{tabular}{c|c|c}
\hline
& Procedure & Average Time\\
\hline
\multirow{3}{2em}{Client} & tracking & 22.5 ms\\
                          & local mapping & 134.1 ms\\
                          & sending map updates & 2.1 ms\\
\hline
\multirow{4}{2em}{Server} & updating global map & 24.8 ms\\
                          & retrieving nearby landmarks & 1.7 ms\\
                          & map merging & 0.60 s \\
                          & pose graph optimization & 1.12 s\\
\hline
\end{tabular}
\end{center}
\end{table}

\begin{figure}[!ht]
  \centering
  \includegraphics[width=3.4in]{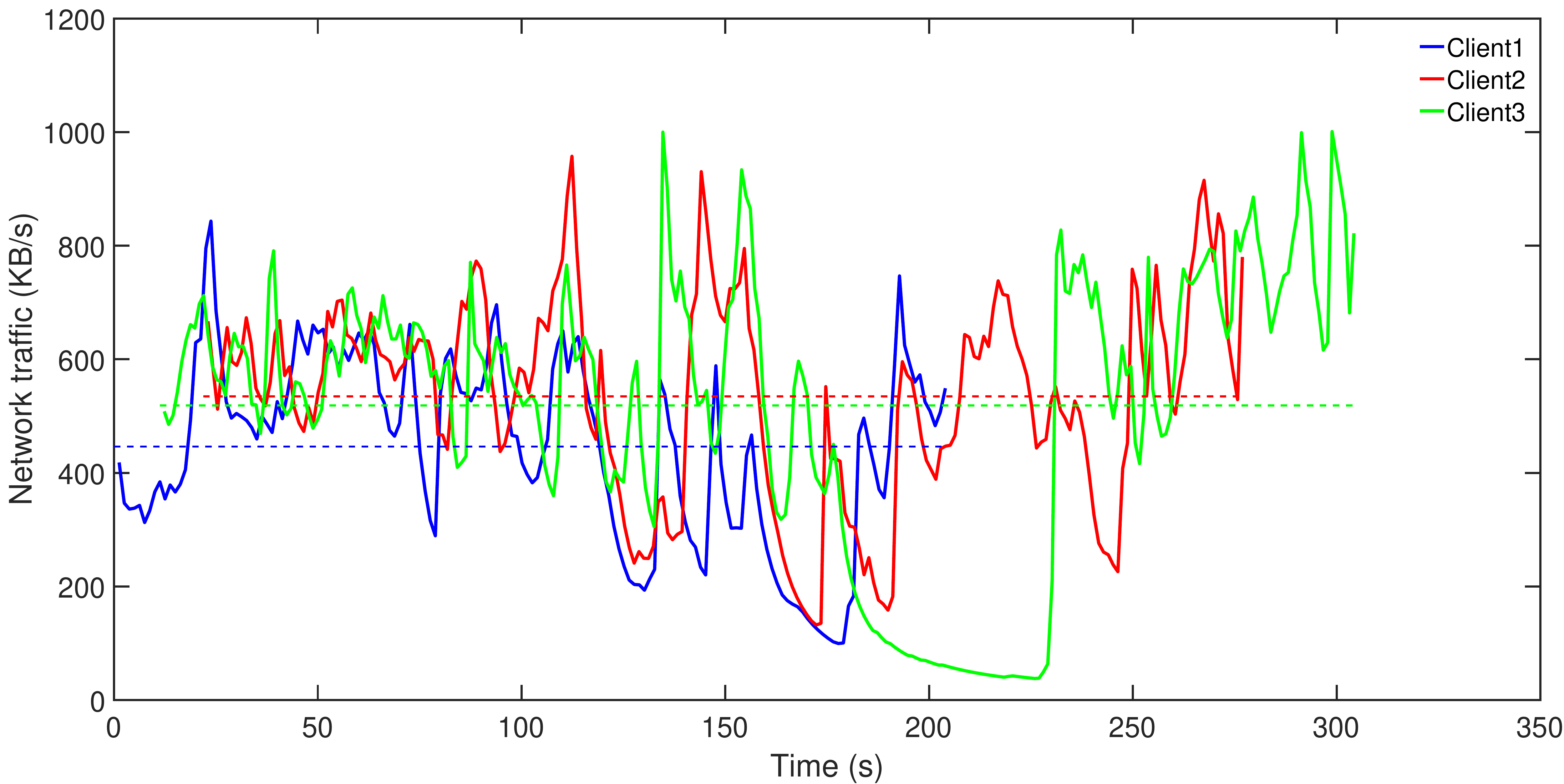}
  \vspace{0.1in}
  {\footnotesize (a) Uplink from each client to the server}
  \includegraphics[width=3.4in]{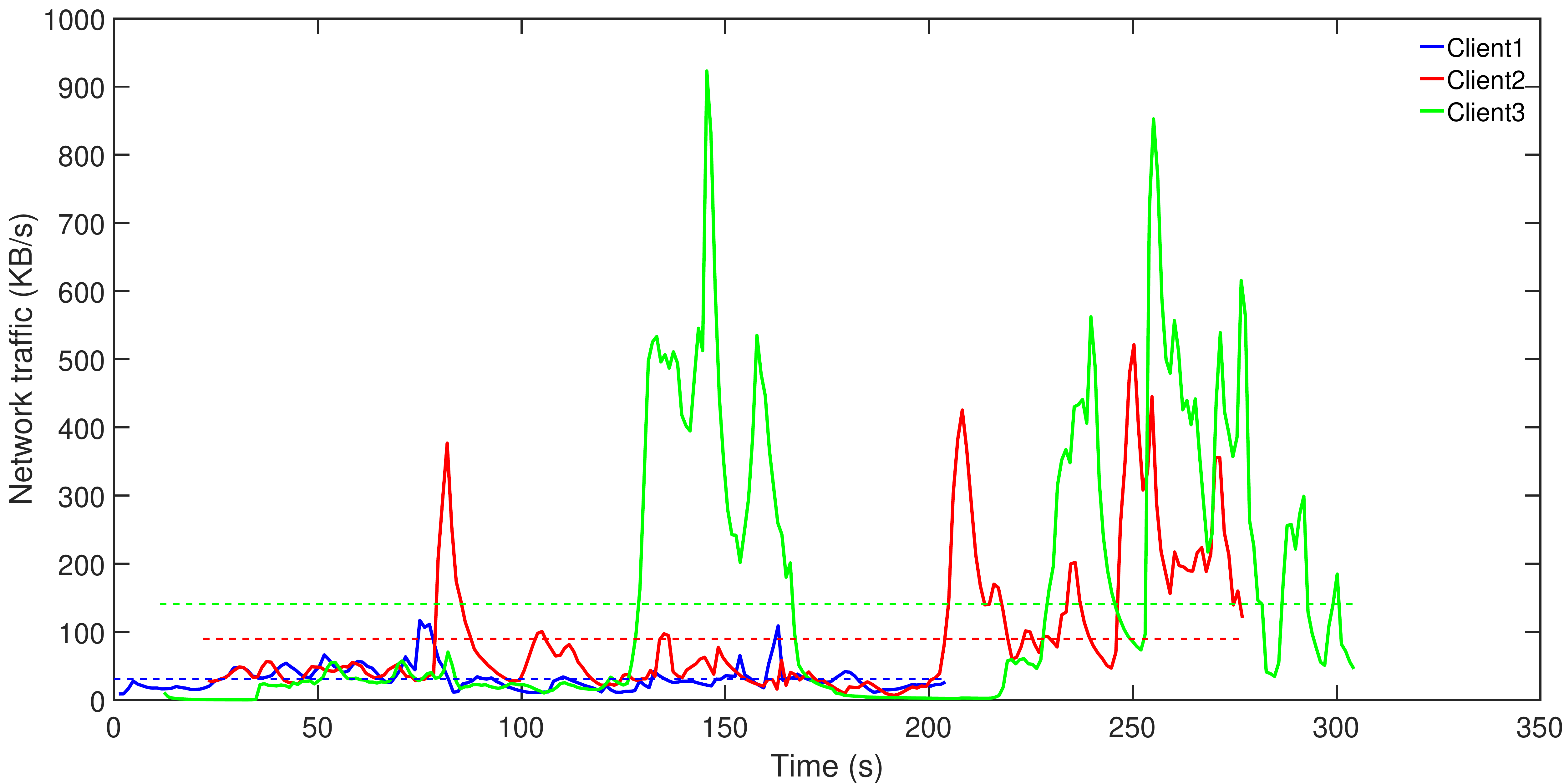}
  {\footnotesize (b) Downlink from the server to each client }
  \caption{The communication cost of three clients running with the three market sequences from the OpenLORIS-Scene datasets. The dotted lines show the average traffic of each client.}
  \label{fig_bandwidth}
\end{figure}

\subsection{Experiments with a Real Multi-camera Robot}

To further verify the system, we test it with a real robot with two RealSense D455 RGB-D cameras. The two cameras are facing front and rear respectively, with no view overlap. Their relative pose is obtained by each calibrating with a third camera on a robot arm on this robot that can move either to the front or to the rear. The robot operates four sessions on a large office floor, exploring different areas in each session, while two clients run on-board with the two cameras. An additional session is made to verify the system's capability with monocular cameras, where only the RGB stream of the front camera is used. The final server map after the five sessions is shown in Fig.~\ref{fig_ryc}. All the visited areas in the five sessions have been mapped together, even that occasionally one client lost when facing a white wall. Most parts of the final map fit well the floorplan of the office.



\subsection{Computation and Communication Cost}

We test the run-time performance of the system with three clients running with the {\verb market } data sequences. The time costs are shown in Table~\ref{tab_time}. All the numbers are averaged with 3 runs on an Intel NUC mini-PC with Intel Core i7-8809G. The costs of clients are moderate, able to support 30 FPS RGB-D input. The costs on the server for each client are pretty low, as map updating and landmark retrieval occur only for each new keyframe, and map merging and pose graph optimization are even less frequent. The proposed landmark retrieval method has a negligible cost of 1.7 ms.

The network traffic loads between each client and the server are shown in Fig.~\ref{fig_bandwidth}. The average traffic for each client is around 0.5 MB/s for uplink and 0.1 MB/s for downlink. As a reference, the average per-agent traffic of CCM-SLAM is 0.37 MB/s for uplink and 0.52 MB/s for downlink \cite{schmuck2019ccm}. They are tested with different data, so cannot be directly compared. Nevertheless, it can be seen that our system has an efficient downlink communication, verifying the effectiveness of our local map augmentation method. We will further optimize the network usage in future works.

\section{Conclusion}

We present a collaborative visual SLAM framework for service robot scenarios. With a client-server architecture and a unified communication protocol, the system is efficient and flexible enough to support multiple robots to collaboratively build and re-use maps. With the novel landmark retrieval method, real-time information sharing is enabled between clients, with affordable computation and communication cost. The system supports heterogeneous cameras and multi-camera robots. We hope that this research can push a step further towards a unified edge server interface to facilitate different service robots to work together.

\bibliographystyle{IEEEtran} 
\bibliography{refs} 
\end{CJK*}
\end{document}